\documentclass{article}

\usepackage{PRIMEarxiv}

\usepackage[utf8]{inputenc} 
\usepackage[T1]{fontenc}    
\usepackage{hyperref}       
\usepackage{url}            
\usepackage{booktabs}       
\usepackage{amsfonts}       
\usepackage{nicefrac}       
\usepackage{microtype}      
\usepackage{lipsum}
\usepackage{fancyhdr}       
\usepackage{graphicx}       
\graphicspath{{media/}}     

\usepackage{cite}
\usepackage{amsmath,amssymb,amsfonts}
\usepackage{algorithmic}
\usepackage{graphicx}
\usepackage{textcomp}
\usepackage{booktabs} 
\usepackage{tabularx} 
\usepackage{xcolor}
\usepackage{ulem}
\usepackage{url}
\usepackage{hyperref}

\usepackage{algorithm}
\usepackage{soul}
\usepackage{amsmath}
\usepackage{subcaption} 
 
\usepackage{amssymb}
\usepackage{pifont}
\title{Motor Focus: Fast Ego-Motion Prediction for Assistive Visual Navigation}

\usepackage[affil-it]{authblk}  

\author[1]{Hao Wang$^{\dag,}$} 
\author[2]{Jiayou Qin$^{\dag,}$} 
\author[1]{Xiwen Chen} 
\author[1]{Ashish Bastola} 
\author[1]{John Suchanek} 
\author[3]{Zihao Gong} 
\author[1]{Abolfazl Razi$^{*,}$} 

\affil[1]{School of Computing, Clemson University} 
\affil[2]{Department of Electrical and Computer Engineering, Stevens Institute of Technology}
\affil[3]{School of Cultural and Social Studies, Tokai University}

\begin{document}

\maketitle

\begin{abstract}

Assistive visual navigation systems for visually impaired individuals have become increasingly popular thanks to the rise of mobile computing. Most of these devices work by translating visual information into voice commands. In complex scenarios where multiple objects are present, it is imperative to prioritize object detection and provide immediate notifications for key entities in specific directions. This brings the need for identifying the observer's motion direction (ego-motion) by merely processing visual information, which is the key contribution of this paper. 
Specifically, we introduce \textbf{Motor Focus}, a lightweight image-based framework that predicts the ego-motion \textemdash the humans' (and humanoid machines') movement intentions based on their visual feeds, while filtering out camera motion without any camera calibration. To this end, we implement an optical flow-based pixel-wise temporal analysis method to compensate for the camera motion with a Gaussian aggregation to smooth out the movement prediction area. 
Subsequently, to evaluate the performance, we collect a dataset including $\mathbf{50}$ clips of pedestrian scenes in 5 different scenarios.
We tested this framework with classical feature detectors such as SIFT and ORB to show the comparison. Our framework demonstrates its superiority in speed ($\mathbf{> 40 FPS}$), accuracy ($\mathbf{MAE = 60 pixels}$), and robustness ($\mathbf{SNR = 23 dB}$), confirming its potential to enhance the usability of vision-based assistive navigation tools in complex environments. The code is publicly available at \url{https://arazi2.github.io/aisends.github.io/project/VisionGPT}.

\end{abstract}


\section{Introduction}
The rapid development of mobile computing has significantly enhanced real-life applications. Technologies including object detection, augmented reality (AR), and assistive visual navigation, have benefited immensely from integrating AI into mobile devices.
As reliance on these digital experiences increases, ensuring user safety through visual assistive technologies has become a critical priority. \cite{kuzdeuov2023chatgpt, al2020smart, li2018vision}.

Motion analysis plays a critical role in assistive visual navigation by utilizing both spatial and temporal information to create a comprehensive visual understanding of dynamic surroundings. 
By analyzing how objects and features move over time within a given space, motion analysis is employed in a suite of technologies to enhance autonomous systems and interactive applications \cite{franke20056d, lee2015moving}.
For instance, visual odometry \cite{scaramuzza2011visual, wang2017deepvo} and ego-motion estimation \cite{monaco2019ego, zhang2020robust, tang2021self}, rely on analyzing consecutive images captured by the camera and extracting features or keypoints that can be tracked across frames, determining the motion of a camera or a vehicle relative to its environment.
Video saliency identifies the most visually conspicuous or attention-grabbing regions within a video sequence, which aims to predict where human observers are likely to focus their attention \cite{tang2018multi, makrigiorgos2019human}. 
Further, optical flow is often used for video stabilization \cite{lim2019real}, object detection \cite{afif2020evaluation, kacorri2017people, bhandari2021object, ashiq2022cnn, bastola2023multi}, image segmentation \cite{yuan2023dynamic}, and depth prediction \cite{zhao2022unsupervised, guizilini2022learning}. These technologies collectively augment this framework, allowing the system to effectively parse and interact with its surroundings by identifying and categorizing environmental elements, thereby ensuring safe and effective navigation.
Furthermore, in the domains of healthcare and sports, motion analysis helps in monitoring human activities and movements, offering valuable insights for training, rehabilitation, and overall well-being assessments \cite{southgate2016motion}. By effectively analyzing spatial and temporal data, motion analysis not only enhances environmental awareness but also significantly improves the interaction capabilities of various technology applications in real-world settings.

\begin{figure}[ht]
    \centering
    \centerline{\includegraphics[width=0.8\linewidth]{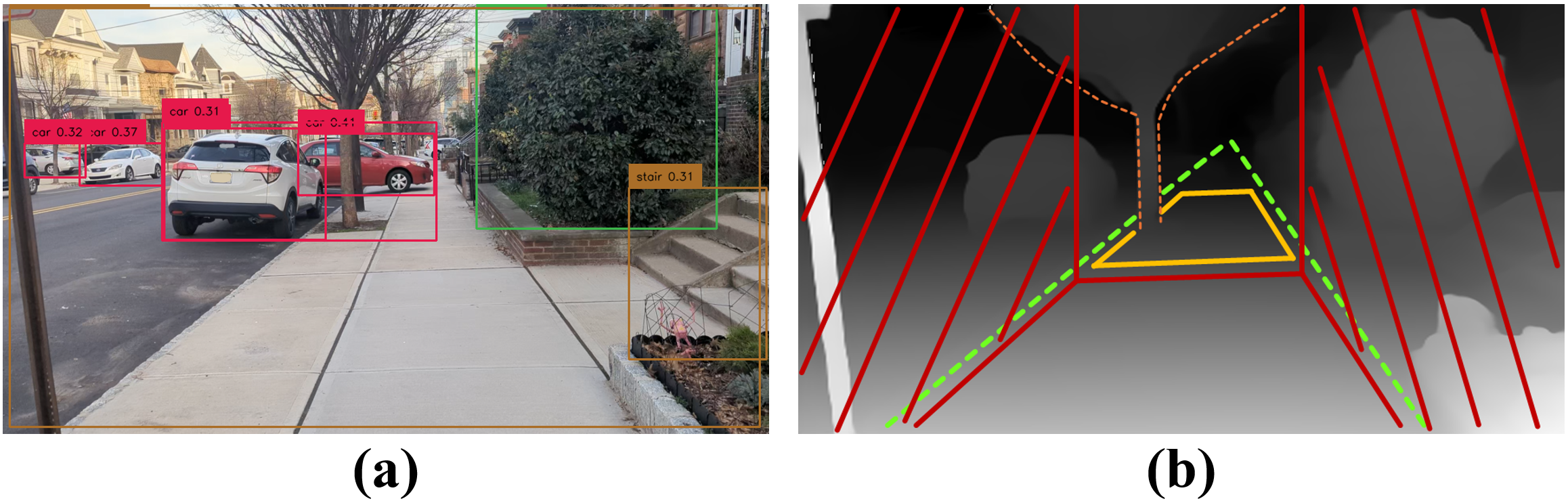}}
    \caption{Concept of Assistive Visual Navigation}
    \label{fig:VIN}
\end{figure}
 
While motion analysis methods have been extensively developed for autonomous driving, their direct implant into mobile devices encounters significant challenges. 
Firstly, although mobile computing capabilities have rapidly advanced, the sensors integrated into mobile devices are typically less sophisticated and more constrained compared to those used in autonomous vehicles. For instance, autonomous vehicles are equipped with a range of high-end sensors such as LIDAR, radar, and multiple cameras that provide detailed, 360-degree environmental feedback \cite{liao2022kitti}. In contrast, mobile devices generally rely on more basic components such as monocular cameras and inertial sensors, which offer less comprehensive data. 
Additionally, autonomous systems are designed to operate continuously and consume considerable power, which is feasible in a vehicle with a substantial power supply. However, for mobile devices, preserving battery life is critical, necessitating energy-efficient solutions \cite{wang2024energy}. The extensive processing power and energy requirements of autonomous driving technologies are thus incompatible with the limited capabilities and energy constraints of typical consumer mobile devices. Consequently, there is a pressing need for cost-effective, energy-efficient motion analysis solutions that benefit the hardware and energy limitations of mobile devices, ensuring that users can enjoy enhanced sensing capabilities without compromising device performance or battery life.
Additionally, human users typically carry devices that are constrained by size, weight, and computational capabilities, limiting the use of intensive real-time processing methods such as Simultaneous Localization and Mapping (SLAM) that are standard in autonomous systems \cite{usenko2015reconstructing}.

Recognizing these limitations, there is a pressing need to develop a navigation aid from the perspective of pedestrians. 
While numerous studies have focused on visual attention, movement analysis is also vital, as the visual focus and body movement can often diverge significantly \cite{tamaru20223d}. Notably, works in visual-based movement prediction in point-of-view camera settings are heavily missing, as vehicle-based research mostly focuses on visual odometry functionality on ego-location recording, while human-based research more focuses on visual attention and head direction prediction \cite{makrigiorgos2019human, liu2020forecasting}.

\begin{figure}[htbp]
    \centering
    \centerline{\includegraphics[width=0.8\linewidth]{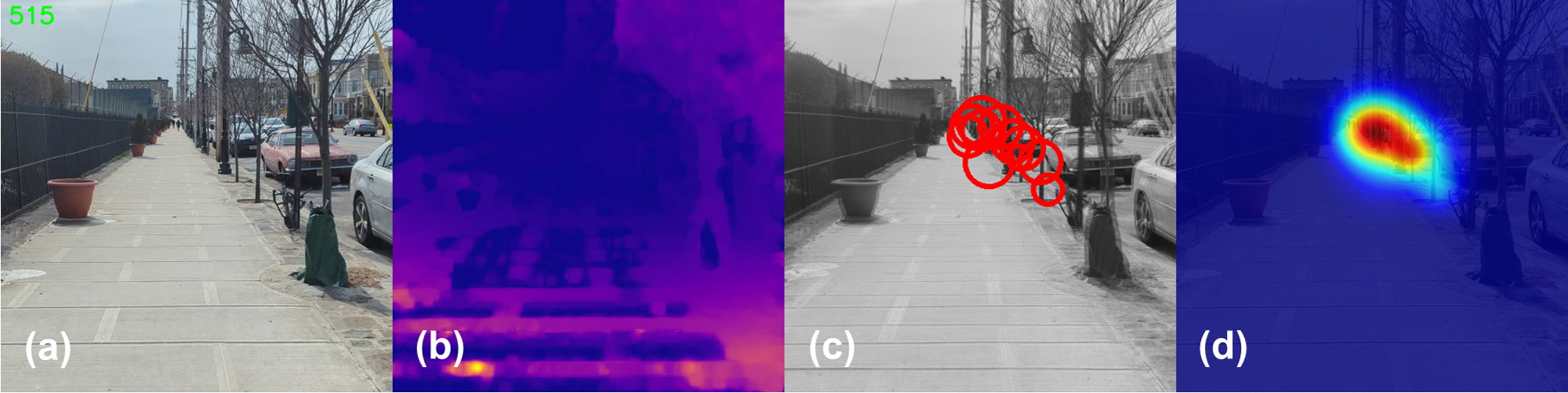}}
    \caption{Motor focus visualization, (a) is the raw RGB image, (b) is the compensated optical flow map, (c) shows the identified attention points of 10 consecutive frames, (d) is the attention map aggregated by the Gaussian distributions of attention points from (c). }
    \label{fig:attention}
\end{figure}

Human movement intention analysis can be useful across a wide array of applications including augmented reality (AR), visual navigation, photography, video stabilization, action cameras (such as 360-degree cameras), drones, and robotics \cite{nguyen2018your}. These fields all benefit from understanding and predicting user movement and orientation, which can enhance interaction quality and system responsiveness. 
For instance, in AR and visual navigation, accurately predicting a user’s movement can improve the alignment and relevance of augmented content, making the experience more intuitive and immersive. In the context of photography and video, understanding motor focus can lead to more effective stabilization techniques and dynamic framing methods. Similarly, in robotics and drone operations, anticipating human motion can enhance collaborative and interactive capabilities. This paper aims to explore motor focus extensively to fill this research void, proposing new methodologies and applications that leverage this underexplored aspect of human-machine interaction.

To bridge this gap, we present \textbf{Motor Focus}, a novel framework for predicting how users physically move and orient themselves in space. 
Specifically, we introduce an optical flow-based pixel-wise temporal analysis that can predict the movement direction of users and simultaneously filter out the unintended and noise-like 
camera motion without any camera calibration. 
We also combined the Gaussian aggregation method to smooth out the projected movement attention area to address the camera shake issue in pedestrian applications. 
Then, to obtain the transform matrix from two consecutive frames, we apply Singular Vector Decomposition (SVD) instead of classical feature mapping to reduce the computation load. Finally, we validate the proposed framework using our self-collected visual navigation-oriented dataset.

\section{Methodology}
\label{sec: method}

This study employs a comprehensive video processing framework designed to enhance visual navigation by analyzing and visualizing motion dynamics in video sequences. The methodology encompasses several key phases, each crucial for extracting meaningful motion data from video frames to aid in navigation tasks. 
The code is publicly available at \url{https://arazi2.github.io/aisends.github.io/project/VisionGPT}.

\begin{figure*}[htbp]
    \centering
    \centerline{\includegraphics[width=1\textwidth]{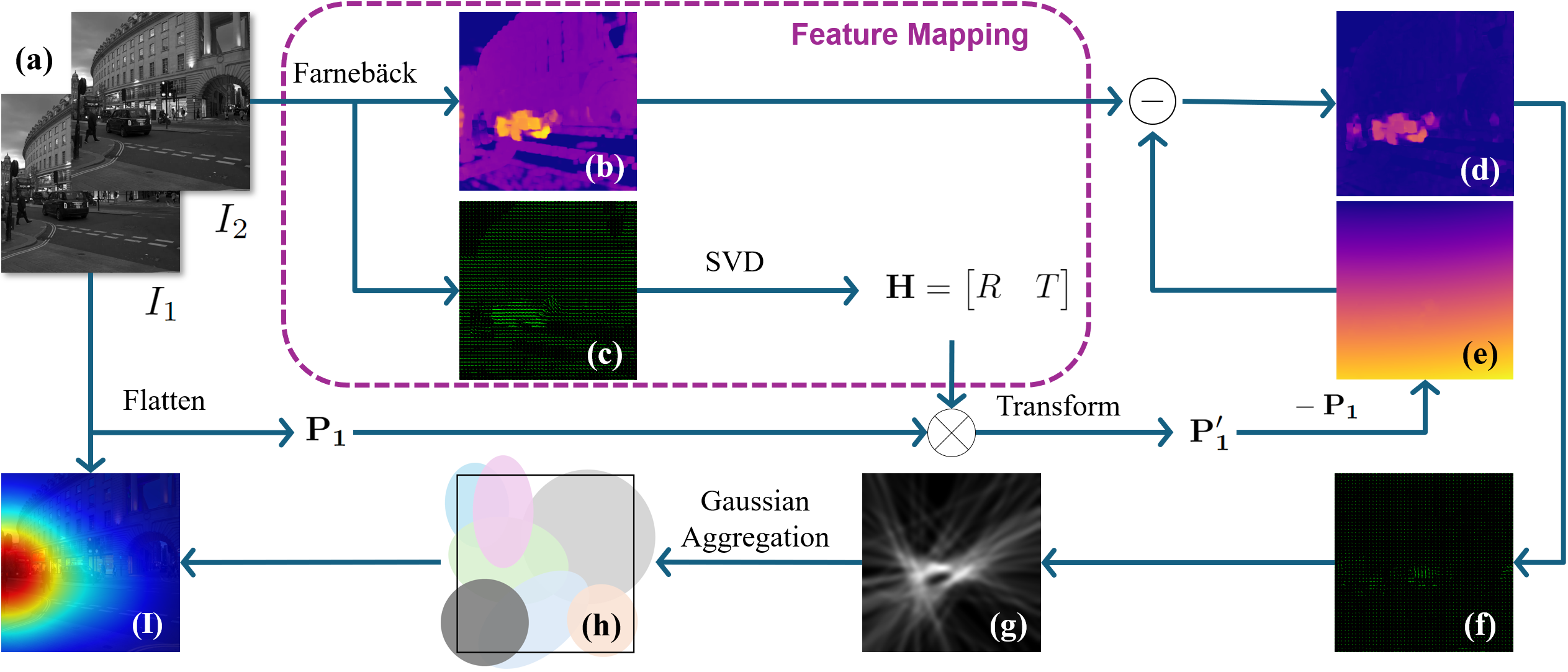}}
    \caption{The proposed framework, (a) is a two consecutive frame pair, (b) is the original optical flow map (magnitude), (c) is the original optical flow field (vector), (d) is the compensated optical flow map, (e) is the camera motion $\epsilon$, (f) is the compensated optical flow field, (g) is the probability map of attention point for $I_2$, (h) is the aggregated gaussian distribution of attention points from (g), and (i) is the attention map for motor focus of frame $I_2$. }
    \label{fig:framework}
\end{figure*}

\subsection{Pre-Processing}

The video capture process is initiated with continuous frame reading from a video stream. Initial frames undergo preprocessing, where they are resized and converted to grayscale to facilitate further analysis. The Scale-Invariant Feature Transform (SIFT) algorithm is implemented early to detect and compute key points and descriptors in the initial frame, establishing a baseline for subsequent frames.
The optical flow, specifically the motion between matched points across frames, is calculated using Farneback's method to provide a dense flow field, highlighting the texture and movement patterns within the scene. 

Specifically, the optical flow field \( \mathbf{f} \in \mathbb{R}^{H \times W \times 2 } \) is computed from two consecutive frames \( I_1 \in \mathbb{R}^{H \times W }\) and \( I_2 \in \mathbb{R}^{H \times W }\) using classical Farneback's \cite{farneback2003two} method:
\begin{align}
\mathbf{f} = \text{Farneback}(I_1, I_2) 
\end{align}

where \( \mathbf{f}_{x_i,y_i} = (dx_i, dy_i) \) represents the displacement vector at position \( (x_i,y_i) \) in image \( I_1 \) that aligns with \( I_2 \). 

The optical flow field \( \mathbf{f} \) between two consecutive images \( I_1 \) and \( I_2 \) provides a dense vector field where each vector points from a pixel in \( I_1 \) to its corresponding location in \( I_2 \), assuming brightness constancy and spatial coherence. This field inherently includes motion from both moving objects and the camera's own motion (ego-motion), which we use the noise term $\mathbf{\epsilon}$ to represent in the rest of the paper.

\subsection{Ego-Motion Estimation using SVD}

Classical image homography estimation such as RANSEC requires a set of matched keypoints $\mathbf{P}_1 $ and $\mathbf{P}_2$ through feature detectors (e.g., SIFT, ORB) to extract the transform matrix that describes the image transform relationships between images \( I_1 \) and \( I_2 \).

In our proposed framework, we use all pixels directly from the gray-scale image \( I_1 \in \mathbb{R}^{H \times W}\).
Specifically, \( \mathbf{P}_1 \in \mathbb{R}^{(H \times W) \times 2}\) is directly mapped from \( I_1 \), denoted as \( \mathbf{P}_1 = \{ \mathbf{p}_1^1, \mathbf{p}_2^1, \ldots, \mathbf{p}_N^1 \} \), where \( N \) is the number of keypoints and is equal to the total pixel number of $I_1$.
For each keypoint \( \mathbf{p}_i^1 = (x_i^1, y_i^1) \) in \( \mathbf{P}_1 \), we find its corresponding point in \( I_2 \) using the displacement \( (dx_i, dy_i) \) obtained from the optical flow field \( \mathbf{f} \) from the previous step:
\begin{align}
\mathbf{p}_i^2 = (x_i^1 + dx_i^1, y_i^1 + dy_i^1), i = 1, 2, \ldots, N
\end{align}

This process ensures that \( \mathbf{P}_2 \in \mathbb{R}^{(H \times W) \times 2}\) contains keypoints that are spatially aligned with \( \mathbf{P}_1 \) in \( I_2 \).

Singular Value Decomposition (SVD) is then used to compute an optimal rigid transformation (rotation \( R \) and translation \( T \)) that best aligns these points \cite{wang2022fast}. Specifically, the cross-covariance matrix \( M \) is computed:
\begin{align}
M = \mathbf{P}_2^T \mathbf{P}_1 = U \Sigma V^T
\end{align}

The rotation matrix \( R \) and the translation vector \( T \) are derived from the SVD components to form the transformation matrix \( \mathbf{H} \):
\begin{align}
\quad R = UV^T, \quad T = \text{mean}(\mathbf{P}_2) - R \times \text{mean}(\mathbf{P}_1) 
\end{align}

Where $\text{mean}(\mathbf{P}_2)$ and $\text{mean}(\mathbf{P}_2)$ are the centroid of each keypoints set.
The transformation matrix \( \mathbf{H} \) is then assembled as:
\begin{align}
\mathbf{H} = \begin{bmatrix} R & T  \end{bmatrix}
\end{align}

\subsection{Ego-Motion Compensation}

The transformation matrix \( \mathbf{H} \in \mathbb{R}^{2 \times 3}\) is then applied to a grid of pixel coordinates \( \mathbf{P_1} \) from \( I_1 \), representing the original positions. The grid modified by the flow \( \mathbf{f} \) gives the new positions \( \mathbf{P_1} \).

Applying \( \mathbf{H} \) to \( \mathbf{P_1} \) provides a prediction of where each grid point would be if only the camera's motion (ego-motion) affected it:
\begin{align}
\mathbf{P_1'} = \mathbf{P_1} \times \mathbf{R} + \mathbf{T} 
\end{align}

The residual term $\mathbf{\epsilon} \in \mathbb{R}^{H \times W \times 2} $ of this study is defined by the difference between the predicted positions of the pixels due to the camera motion and the predicted positions of the pixels without camera motion, which represents the displacement caused exclusively by the camera's motion, ignoring any independent object movements:
\begin{align}
\epsilon = reshape(\mathbf{P_1'} - \mathbf{P_1})
\end{align}

The compensated optical flow then can be derived by correcting the raw optical flow \( \mathbf{f} \) for the motion attributable to the observer's (camera's) own campaign. This correction ensures that \( \mathbf{f}' \) predominantly reflects the motion of objects relative to the observer, rather than due to the observer's motion. This is expressed mathematically as:
\begin{align}
\label{eq:camera}
\mathbf{f}' = \mathbf{f} - \mathbf{\epsilon}
\end{align}

Here, \( \mathbf{f}' \) represents the motion vectors corrected for camera motion, highlighting only those movements that are due to objects moving in the scene rather than the camera itself.

\subsection{Project Movement Direction }

Given the initial set of vectors \( \mathbf{v}_j = \mathbf{p}_1^j + \mathbf{f}_j' \), assuming that each \( \mathbf{v}_j \) potentially indicates a trajectory toward a potential focus point, the goal is to find a point \( \mathbf{c} \) that most of these vectors converge toward:
\begin{align}
\min_{\mathbf{c}} \sum_j \| \mathbf{c} - \text{proj}_{\mathbf{v}_j}(\mathbf{c}) \|^2 
\end{align}
However, for complex scenes, there might be multiple focuses, which shift the point \( \mathbf{c} \) from the optimum location.
To measure how well a point \( \mathbf{c}_k \) (a candidate point from the k-th cluster) aligns with the vectors \( \mathbf{v}_j \), we formulate an optimization problem to minimize these deviations for each cluster. Specifically, for each candidate focus \( \mathbf{c}_k \):
   \begin{align}
\mathbf{c}_k = \sum_{j=1}^{M} \frac{\left\| \mathbf{c}_k - \text{proj}_{\mathbf{v}_j}(\mathbf{c}_k) \right\|^2}{M}, {j \in \text{cluster}_k} 
   \end{align}

Where \( \text{proj}_{\mathbf{v}_j}(\mathbf{c}_k) \) is the projection of \( \mathbf{c}_k \) onto the line defined by \( \mathbf{v}_j \).

Then we calculate the score of each candidate focus \( \mathbf{c}_k \):
\begin{align}
\text{Score}(\mathbf{c}_k) = \sum_{j=1}^{M} j, {j \in \text{cluster}_k}
\end{align}

Select the \( \mathbf{c}_k \) with the highest score, which indicates the maximum number of vectors \( \mathbf{v}_j \) converging towards it.

\subsection{Attention Area Smoothing}
To stabilize the movement attention area across frames and reduce the effect of transient shaking or focus shifts, the Gaussian aggregation method is applied at each frame's focus point $\mathbf{c}_i$ to create a smooth attention mask over multiple frames. The spread of each Gaussian is determined by the inverse of the mean magnitude of the optical flow field, $\mathbf{U}_i$, representing the activity level or movement intensity in the frame. The formulation can be described as follows: 
\begin{align}
K_{\text{new}}(x, y) = \sum_{i=1}^{n} K_i(x,y|\mathbf{c}_i, \sigma_i),~\quad
\sigma_i \propto \frac{1}{U_i}, \\
U_i = \sqrt{u_1^2 + u_2^2 + ... + u_N^2} 
\end{align}

Here, \(K_i(x,y|\mathbf{c}_i, \sigma_i)\) is the Gaussian distribution from the \(i\)-th frame with its standard deviation \(\sigma_i\), and \(n\) is the number of distributions (frames) considered.

The final Gaussian aggregation mask effectively smoothies and stabilizes the attention area by filtering out the ambiguous focus points due to the strong camera motion, as shown in Figure \ref{fig:attention}

\section{Experiments}

In this section, we evaluate the proposed method both qualitatively and quantitatively.

\subsection{Data Collection}

To test the performance of our framework, we collected a dataset that is specialized for visual navigation.  In advance, each video clip is observed by three researchers frame by frame, and a pixel location $(x,y)$ of moving direction is annotated for each frame. 

\begin{table*}[htbp]  
    \centering      
        \resizebox{0.6\linewidth}{!}{
    \begin{tabular}{cccccc} 
        \toprule      
        Location&Scene  &Movement &Weather& Clips&Total length\\\midrule
 Urban  &Sidewalk&Scooter &Cloudy&   8&10 mins\\
 Suburban& Bikeline& Scooter & Cloudy& 5& 6 mins\\ 
 Urban &Park&Scooter&Cloudy&   6&5 mins\\
 City &Road& Biking&Sunny&   5&5 mins\\
 City& Sidewalk& Biking& Sunny& 7& 6 mins\\
 City&Park& Biking&Cloudy&   5&5 mins\\
 Town&Park& Walking&Cloudy&    6&4 mins\\
 Town & Sidewalk& Walking & Sunny& 8& 7 mins\\
 City& Coast& Walking & Sunny& 2& 5 mins\\
        Suburban&Theme Park& Walking &Rain&   3&6 mins\\ \bottomrule
    \end{tabular}}
    \caption{Action type and length of collected dataset.}
        \label{tab:data}
\end{table*}

\begin{figure}[ht]
    \centering
    \centerline{\includegraphics[width=0.8\linewidth]{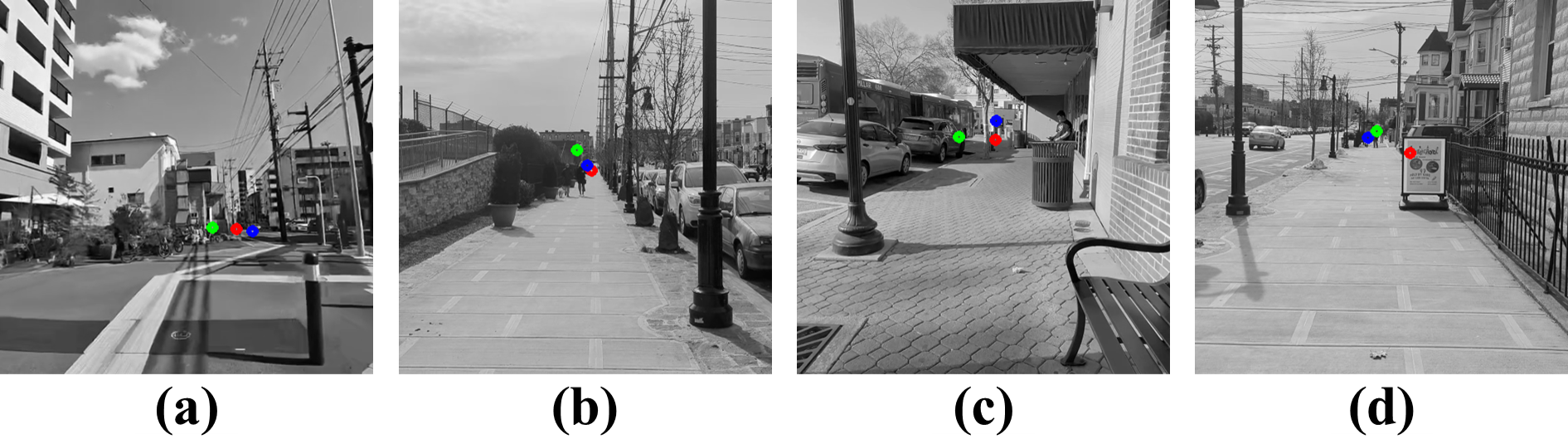}}
    \caption{The samples of the collected dataset.}
    \label{fig:dataset}
\end{figure}

Figure \ref{fig:dataset} shows the sample of the collected dataset, where colored points represent the annotation from different researchers. 
The ground truth is calculated by the average of three different pixel locations. 
Specifically, Figure \ref{fig:dataset} (a) is a biking scene, (b) and (d) is a scooter riding scene, and (c) is a walking scene.

\subsection{Camera-Motion Compensation}
To evaluate the compensation performance of the proposed method, we compare the vanilla dense-optical flow and the proposed compensated optical flow in a series of scenes to showcase its capability to capture camera motion.

\begin{figure*}[htbp]
    \centering
    \centerline{\includegraphics[width=1\textwidth]{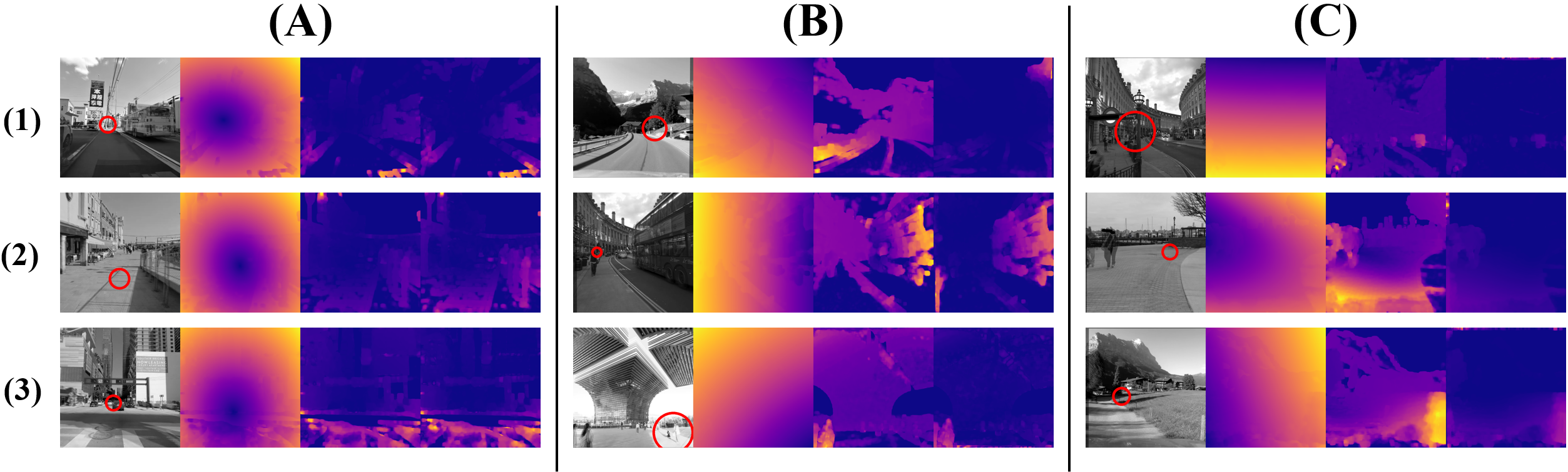}}
    \caption{Visualization of ego-motion compensation, each image consists of four cells, from left to right: grayscale image with predicted moving direction, the magnitude of camera motion $\epsilon$ (ego-motion), raw optical flow (vanilla dense-optical flow), and optical flow with ego-motion compensation.}
    \label{fig:grid}
    \vspace{-0.1 in}
\end{figure*}

As shown in Figure \ref{fig:grid}, the compensated optical flow can filter the motion caused by camera shifting, distinguishing the relatively moving objects. Meanwhile, nearby objects' moving speed and direction can also be estimated from the compensated optical flow map. 

More importantly, Figure \ref{fig:grid} indicates that the visual focus can differ greatly from the motor focus. For instance, in Figure \ref{fig:grid} group B, the compensated optical flow and camera motion indicated the camera moving potential is different from the actual body movement. In B1 and B3, the camera moves toward the left corner, while the user moves toward the right side. In B2, both the camera and the user move toward the left. 
In Figure \ref{fig:grid} group C, the camera in both C2 and C3 are turning right, while the user in C2 is moving right and the user in C3 is moving left. In C1, the user stands statically, while the camera motion suggests that the camera is slowly pitching up. All results are proved in the actual video footprints.

Interestingly, when the camera moving direction and the user movement are aligned straightforwardly, the vanilla optical flow and the compensated optical flow become identical, and the camera motion map tends to overlap with the motor focus area, as shown in Figure \ref{fig:grid} group A.

\subsection{Ego-Motion Prediction}

We applied the proposed method to predict the center of moving direction and we used Mean Absolute Error (MAE), and Mean Squared Error (MSE) to compare the predicted location $(\hat{x},\hat{y})$ with the annotated ground truth$(x,y)$. Furthermore, we used the Signal-to-Noise Ratio (SNR) to compare the vertical motion $y_i \in \mathbf{c_i}$ along the time in a selected scene.

\begin{figure}[ht]
    \centering
    \centerline{\includegraphics[width=0.8\linewidth]{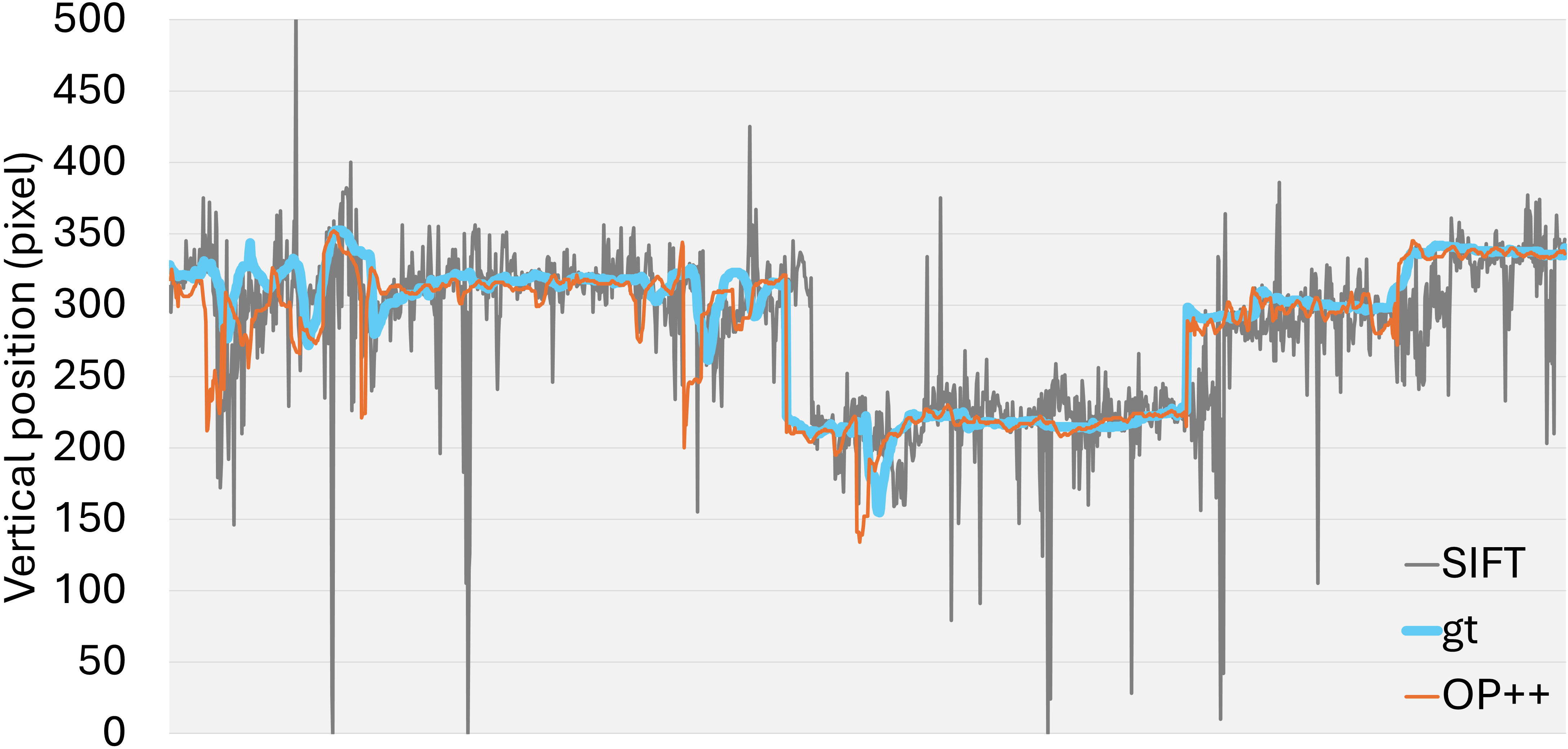}}
    \caption{Plot of the focus points (vertical position $y_i \in \mathbf{c_i}$) on the selected scene.}
    \label{fig:snr}
\end{figure}

Our framework seamlessly works with other feature detectors, such as Scale-Invariant Feature Transform (\textbf{SIFT}) and Oriented FAST and Rotated BRIEF (\textbf{ORB}), and optical flow methods such as the Lucas-Kanade (\textbf{LK}) method, by simply replacing the Feature Mapping module of the proposed framework, as shown in Figure \ref{fig:framework}. We show the comparison between these methods in Table \ref{tab:time}, where $\mathbf{OP}$ represents the vanilla dense-optical flow method (Farneback), $\mathbf{OP+}$ represents the $\mathbf{OP}$ method with camera motion compensation, and $\mathbf{OP++}$ represents compensated $\mathbf{OP}$ with the Gaussian aggregation. All frames are resized to $512 \times 512$ for experiments.

\begin{table}[htbp]  
    \centering      
    \caption{Performance comparison}
    \resizebox{0.8\linewidth}{!}{
    \begin{tabular}{>{\centering\arraybackslash}p{0.15\linewidth}>{\centering\arraybackslash}p{0.15\linewidth}>{\centering\arraybackslash}p{0.15\linewidth}cccc} 
        \toprule      
        Feature Detection Method&Matching Time (ms)&Total Time (ms) &FPS&MAE &MSE (x1000)&SNR (dB)\\\midrule
 LK&4.86&27.60& 36.24& 103.89& 11.88&16.47 \\
  ORB & 5.34&26.76& 37.36& 112.35& 11.95&16.44 \\
   SIFT& 35.49&58.31& 17.15& 93.34& 8.40&18.45 \\
 OP& 0.91& 19.38& 51.59& 108.21& 11.37&14.27 \\
 OP+& 0.91& 22.49& 44.47& 90.17& 8.82&19.45 \\
 \textbf{OP++}& \textbf{0.91}&23.45& 42.64& \textbf{60.66}& \textbf{4.26}&\textbf{23.09 }\\\bottomrule
    \end{tabular}}
    \label{tab:time}             
\end{table}

Notably, the matching time ($<1 ms$) and FPS ($>40$) of all three $\mathbf{OP}$ methods stand out from other classical methods as we directly use the optical flow field $\mathbf{f}$ for feature mapping. By taking advantage of SVD, the linear operation processes a total of $\mathbf{262,144}$ elements (all pixels) without additional computation cost. 
Meanwhile, the camera motion compensation helps to approach lower predicting error in $\mathbf{OP{+}}$ compared to vanilla optical flow ($\mathbf{OP}$). 
Furthermore, the proposed $\mathbf{OP++}$ method achieves even better results. Specifically, it obtained the lowest prediction error of motor focus in both MAE and MSE. More importantly, the Gaussian aggregation helps smooth the prediction especially the vertical location of the motor focus point $\mathbf{c_i}$. 
Figure \ref{fig:snr} confirms the accuracy (better alignment with ground truth) and stability (fewer fluctuations) of our method ($\mathbf{OP++}$ ) compared with the SIFT.

\section{Conclusion}
This study presents the \textbf{Motor Focus} --- a novel image-based framework for motion analysis, specifically designed for predicting ego motion for pedestrians. By utilizing camera-motion compensation with Gaussian aggregation, our approach effectively tackles the camera shaking challenge, enhancing movement prediction accuracy and stability. Our experimental results, both qualitative and quantitative, validate the superiority of our method over classical feature detectors, especially in accuracy and computation cost. 
Our method can be utilized by blind assistive navigation tools to prioritize notifications based on the detected objects' alignment with ego-motion direction.

\bibliography{references} 
\bibliographystyle{unsrt}

\end{document}